# IGAN: Inferent and Generative Adversarial Networks

**Dr. Luc VIGNAUD**
ONERA, the French aerospace lab,
FRANCE

Luc.vignaud@onera.fr

## ABSTRACT

*I present IGAN (Inferent Generative Adversarial Networks), a neural architecture that learns both a generative and an inference model on a complex high dimensional data distribution, i.e. a bidirectional mapping between data samples and a simpler low-dimensional latent space. It extends the traditional GAN framework with inference by rewriting the adversarial strategy in both the image and the latent space with an entangled game between data-latent encoded posteriors and priors. It brings a measurable stability and convergence to the classical GAN scheme, while keeping its generative quality and remaining simple and frugal in order to run on a lab PC. IGAN fosters the encoded latents to span the full prior space: this enables the exploitation of an enlarged and self-organised latent space in an unsupervised manner. An analysis of previously published articles sets the theoretical ground for the proposed algorithm. A qualitative demonstration of potential applications like self-supervision or multi-modal data translation is given on common image datasets including SAR and optical imagery.*

## 1. INTRODUCTION

### 1.1 Background on Generative Adversarial Networks and Variational Auto Encoders

Generative Adversarial Networks (GANs) [1] and Variational Auto Encoders (VAEs) [2] are standard methods for learning generative models of complex data distributions. In their basic form, GANs tend to generate better quality data samples than VAEs, but suffer from stability and mode collapse problems [3]. GANs were not originally built to produce inference unlike VAEs. Inference is however crucial for numerous tasks that are much easier when processed in the latent space rather than in the data space: e.g. classification, clustering, disentanglement, interpolation or extrapolation, compression, domain and style transfer, to name a few [4] [5][6][7].

Let $X_r$ be a real data distribution and $Z_p$ a known prior model, e.g. a vector of Gaussians. Let $\tilde{X}_p$ be a data distribution generated from $Z_p$, and $\tilde{Z}_r$ a latent distribution estimated from $X_r$. $E$ is defined as an Encoder that encodes data to latents (e.g. $\tilde{Z}_r = E(X_r)$), $G$ a Generator that generates data from latents (e.g. $\tilde{X}_p = G(Z_p)$ ), and $D$ a Discriminator with a single decision output. When needed, the notation $D_x$ and $D_z$ will be used to separate a Discriminator working respectively on data or latents (e.g. $D_x(X_r)$ or $D_z(\tilde{Z}_r)$).

A Vanilla GAN is composed by a Generator and a data Discriminator. It builds an adversarial game between real and generated (*fake*) data, using $D_x(X_r)$ and $D_x(\tilde{X}_p) = D_x\left(G(Z_p)\right)$ to minimize their Jensen Shannon Divergence [1]. Such architecture aligns the generated data distribution with the real one, but ignores latents inference.

A Vanilla VAE uses an Encoder and a Generator to minimize the two terms of the ELBO (Evidence Lower Bound Optimization) [2]. The first term minimizes the Kullback Liebler Divergence between the latents encoded from real data and the prior. The second one is a regularization term that minimizes the reconstruction error in the real data space $\varepsilon(X_r, G(E(X_r)))$. It aligns the posterior distribution of latents encoded from real data with prior, and constraints the reconstruction data error to follow a given distribution (e.g. Gaussian if $\varepsilon = L_2$). However, the Euclidean distance yields blurry data reconstructions, partly because its data similarity judgment is not robust to minor data transformations such as translations.

## 1.2 Bridging the gap between GAN and VAE

Many solutions have been proposed to combine the best of the two worlds: VAEGAN [8], BIGAN [9], ALI [10], ALICE [11], AGE [12], IntroVAE [13], BigBIGAN [14], AEGAN [15], DALI [16], LIA/GAN [17], ALAE [18]… Basically, all these methods bring a latent inference to the classical GAN, and compete to better align distributions in both latent and data spaces, hopefully for both posteriors and priors. Their relative performance is often measured by the ability of the algorithm to reconstruct the input data $X_r$ from the auto-encoded regeneration of its inferred latent: $X_r \sim G(E(X_r))$. It is yet impossible to expect a perfect reconstruction of real data since the dimension space reduction from data to latents implies that some information has to be lost: details or outliers that are not statistically relevant to the training dataset needs be filtered out of the generation process, and consequently out of the reconstruction. Other reconstruction choices have to be made depending on the input data types and the final application of the algorithm. Most methods, if not all, are focused on photographic optical images applications, and thus reconstruct images via an Euclidian distance alternative, for instance with a perceptual loss built on features extracted from the latest layers of an external pretrained classifier, as inspired by StyleGAN [19].

Only a few methods consider the latent space on a par level with the data space, while it is *de facto* where most of the post-training data processing will be applied and most useful applications drawn. Yet, if we follow Rabelais's famous advice to "*break the bone and suck out the substantific marrow*" [*Gargantua*], the reconstruction of prior latents should even prevail on real data reconstruction to assure a bijective inversion between the Generator and the Encoder on the prior space, i.e. to form an optimal <u>latent</u> auto-encoder. In ALI [10], authors already suggested that an Encoder could use such a reconstruction while playing the conventional GAN scheme. Nevertheless, ALI, its twin BIGAN [9] (Bidirectional Generative Adversarial Networks), or its later extension BigBIGAN [14] used a large Discriminator embedding both images and latents whose aim is to equalize the joint probabilities $p(X_r, E(X_r)) = p(G(Z_p), Z_p)$, but with no explicit reconstruction term. Unfortunately, this lack of deterministic point-wise matching between $X_r$ and $G(E(X_r))$ was shown to produce poor data reconstructions. Recently, DALI [16] proposed to guide the GAN adversarial game with a reconstruction term exclusively computed in the latent space to benefit from an easier reconstruction error model than in the data space. Then, two similar techniques (LIA/GAN [17] and ALAE [18]) introduced an intermediate latent space reconstruction, with a complementary latent auto-encoder linked to the prior latent space. AEGAN [15] presents a clever revisit of the famous unpaired image-to-image translator CycleGAN [20], where the latent space takes the place of one of the target image space. The problem is then seen as a double GAN in both image and latent spaces (but not embedded). A cycle consistency loss is built on the reconstruction of auto-encoded latents and images to assure the data-to-latent pairing that needs to be learned.

This analysis suggests that:

- a latent GAN counterpart is mandatory to allow the encoded latents distribution to span the full extension of prior space and explore all realistic possibilities for generating data.

- the latent and data GANs should be embedded within a single Discriminator to emphasize the joint probabilities equalization. This embedding needs to be done on an intermediate space, as already required by their dimension difference.

- a cycle consistency is needed for data-to-latent pairing, and requires at least the auto-encoded latents reconstruction.

With IGAN (Inferent Generative Adversarial Networks), I propose to aggregate this framework with an original entangled and embedded adversarial game in both latent and image space. The architecture is kept fairly light with a low computation and memory load in order to run on a lab PC with a single GPU.

## 2. IGAN: INFERENT GENERATIVE ADVERSARIAL NETWORKS

### 2.1 IGAN architecture

IGAN has been built with the intention to give an equal importance to both data and latent spaces. Since the adversarial game between fake $\tilde{X}_p$ and real data $X_r$ is responsible for the high quality of data generation, the latent generation/inference should also benefit from its own adversarial game between *fake* encoded latents of real data $\tilde{Z}_r$ and *real* prior model $Z_p$. Previous works like DALI [16] and BigBIGAN [14] pointed on the importance to equalize the joint probabilities $p(X_r, E(X_r)) = p(G(Z_p), Z_p)$. I state that it is even better to equalize the following joint probabilities: $p(X_r, Z_p) = p(G(Z_p), E(X_r))$. This modified condition enables us to rewrite the coupled adversarial game in an embedded space where whatever $(X_r, Z_p)$ couple represents a "True data-latent couple", and $(G(Z_p), E(X_r))$ stands for the "False/generated data-latent couple" to play the adversarial game with. The overall IGAN architecture is presented in the following Figure 1.

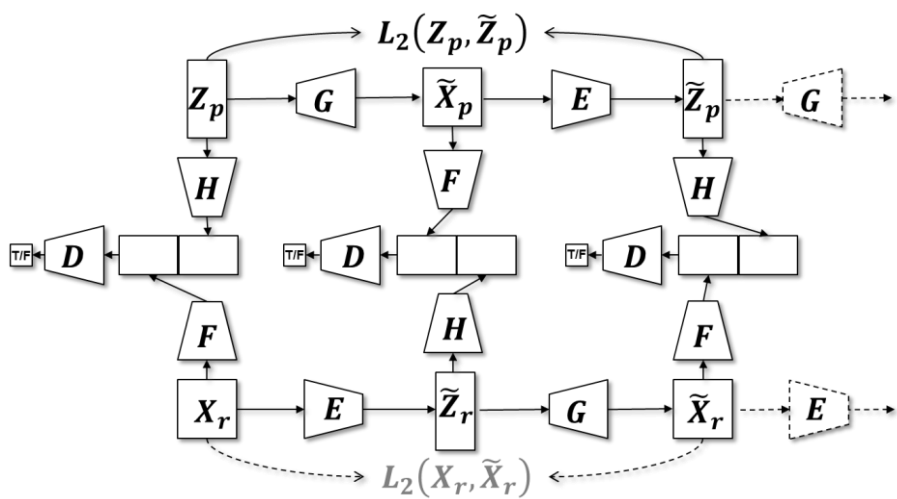

**Figure 1: IGAN: Inferent and Generative Adversarial Networks architecture**

Prior latent $Z_p$ and real data $X_r$ are first encoded into secondary latents using two encoders $H$ and $F$, whose outputs are then concatenated to feed a unique Discriminator $D$. This sets the reference for what will be used as the "True data-latent couple" discrimination. Prior latents are also turned into generated data $\tilde{X}_p$ through Generator $G$, while the real data are encoded into generated latents $\tilde{Z}_r$ by Encoder $E$. These "False/generated data-latent couples" will be driven through the same [$H$ & $F$ encode, concatenate and $D$ discriminate] process as the "True data-latent couples" to feed the adversarial game. This seeds the following adversarial minimax optimization goal for the networks:

$$\min_{G,E} \max_{D,F,H} \left\{ \mathbb{E}_{X_r, Z_p} [D(F(X_r), H(Z_p))] - \mathbb{E}_{X_r, Z_p} [D(F(G(Z_p)), H(E(X_r)))] \right\} \qquad (1)$$

In Figure 1, one can easily spot the $(E \circ G)$ data auto-encoder and $(G \circ E)$ latent auto-encoder, respectively at the bottom and top lines. It is important to see that we will also be able to use the reconstruction of real data and prior latents to follow $p(G(E(X_r)), E(G(Z_p))) = p(X_r, Z_p)$ and, furthermore, the reconstruction of generated data and latents encoded from real data with $p(G(E(G(Z_p))), E(G(E(X_r)))) = p(X_r, Z_p)$. Incorporating the corresponding additional adversarial games in eq. (2) will ensure that not only the reconstructed real and fake data, but also the reconstructed real and fake latents will follow the real data and prior latent statistics (respectively).

The proposed IGAN architecture is not as greedy as it may appear at first sight. It shares the same networks ($G$ and $D \circ F$) as the classical GAN, and only requires the addition of a data Encoder $E$ and a small latent-to-latent network $H$ (moreover $H$ is optional, see Section 2.2). Whereas AEGAN [15] was built on CycleGAN [20] grounds, IGAN may be related to the DiscoGAN [21] architecture with the latent space replacing one of the two image data spaces, and with the major difference that it uses a unique Discriminator instead of the two Discriminators traditionally used in such unpaired image-to-image translators (one per image side).

However, alignment of the latent and image distributions is not sufficient to guarantee the element wise latent or data reconstruction. A cycle consistency loss is then introduced to learn the data-to-latent pairing. Let *fake* data $\tilde{X}_p$ be re-encoded by $E$ to $\tilde{Z}_p = \mathrm{E}\left(\mathrm{G}(Z_p)\right)$ as reconstructed/autoencoded latent priors. Thus, a regularization term $L_k(\tilde{Z}_p, Z_p)$ is added to the Encoder and Generator loss to ensure that the reconstruction errors of encoded latents follow a simple distribution (i.e. Gaussian when using $L_2$ norm). With a sigmoïd activated Discriminator, the losses to be minimized by the different networks translate into eq. (2).

$$\begin{cases} Loss(D,F,H) & = & \begin{aligned} & 3*\mathbb{E}_{X_r,Z_p}[-\log(D(F(X_r),H(Z_p)))] + \mathbb{E}_{X,Z_p}[-\log(1-D(F(G(Z_p)),H(E(X_r))))] \\ & +\mathbb{E}_{X_r,Z_p}[-\log(1-D(F(G(E(X_r))),H(E(G(Z_p)))))] + \mathbb{E}_{X_r,Z_p}[-\log(1-D(F(G(E(G(Z_p)))),H(E(G(E(X_r))))))] \end{aligned} \\ Loss(E,G) & = & \begin{aligned} & \mathbb{E}_{X,Z_p}[-\log(D(F(G(Z_p)),H(E(X_r))))] + \mathbb{E}_{X_r,Z_p}[-\log(D(F(G(E(X_r))),H(E(G(Z_p)))))] \\ & +\mathbb{E}_{X_r,Z_p}[-\log(D(F(G(E(G(Z_p)))),H(E(G(E(X_r))))))] + \propto* \mathbb{E}_{Z_p}\left[\left\|\tilde{Z}_p - Z_p\right\|^2\right] \end{aligned} \end{cases} \quad (2)$$

Where $\propto$ is a hyper-parameter weighting. Additionally, a cycle consistent data reconstruction loss between $X_r$ and $\tilde{X}_r = \mathrm{G}\left(\mathrm{E}(X_r)\right)$ may also be added to help the data-to-latent pairing process. But since the $L_k$ norm is often suboptimal on raw data and may generate noisy reconstructions, an alternative is to minimize the real latent reconstruction error (e.g. $L_k(\mathrm{E}(\mathrm{G}(\mathrm{E}(X_r))), \mathrm{E}(X_r))$, or the difference between some intermediate Encoder layers activations when fed with $X_r$ and $\tilde{X}_r$.

Upon convergence, the Generator and Encoder become invertible relatively to each other when fed with latents complying with prior distribution, or data generated from prior. It is worth noting that the reconstruction of real data will still align with the real data distribution, but will ignore details that are not consistent with the somehow *compressed* distribution of real data learned by IGAN. Nevertheless, the simplified latent distribution extracted from real data will be very useful since it has been self-organised by IGAN to span a maximum of the full prior space. Therefore, *close* data in the data space (in the semantic sense) should be encoded to *close* latents (in $L_2$ sense), and *close* latents should generate *close* data.

## 2.2 Implementation choices:

It is worth noting that the IGAN architecture do not impose a specific layer implementation for the neural networks, and may be tested with the ones previously suggested in [8] to [18], using resnet nodes [14], attention layers, or even the latest transformers like in TransGAN [23].

Since applications will be shown Section 3 on image datasets with mid-size image dimensions (e.g. 64*64 pixels), I used the very classical DCGAN implementation [22] using [Stride 2 Convolution, BatchNorm, Relu] down-sampling blocks for the data Encoder E and F networks, and [Stride 2 Transposed Convolution, BatchNorm, Relu activation] up-sampling blocks for the Generator G. The secondary latent Discriminator is a simple MLP made of a couple of Fully Connected layers. The latent to secondary latent Encoder H is also made of a couple of Fully Connected layers: I experimentally noticed that this network could be omitted and I directly fed primary latents to the concatenated input of the Discriminator.

With conventional GANs, it is often hard to estimate the convergence of the optimization, and stopping criteria remain unclear. With IGAN, multiple performance and convergence indicators are available:

- The Discriminator scores monitor the GAN balance within the latent-data generation process.

- The prior latent reconstruction error (e.g. $L_k\big(\mathrm{E}(\mathrm{G}(Z_p)), Z_p\big)$ ) measures the latent space consistency and Encoder/Generator bijective invertability.

- The *fake* data reconstruction error (e.g. $L_k\big(\mathrm{G}(Z_p), \mathrm{G}(\mathrm{E}(\mathrm{G}(Z_p)))\big)$) may also be used to test the generated data space consistency and Encoder/Generator bijective invertability.

- Real data reconstruction error (e.g. $L_k(X_r, \mathrm{G}(\mathrm{E}(X_r)))$) should be used cautiously since outliers or statistically non-coherent details will be ignored from reconstructed data. The real latent reconstruction error (e.g. $L_k(\mathrm{E}(\mathrm{G}(\mathrm{E}(X_r))), \mathrm{E}(X_r))$ ) may be used as a substitute.

## 3. APPLICATIONS

### 3.1 Data and latent reconstruction

The AFRL SAMPLE database is a recently released collection of simulated Synthetic Aperture Radar (SAR) signatures of ground targets paired with MSTAR image chips [24]. The dataset contains ten targets observed at fixed incidences (15° to 17°) for 360° span of azimuth orientations. IGAN was trained on the real SAR image chips and the reconstruction of both trained and generated images is shown Figure 2. It should be noticed that this training dataset is very small (less than 6000 images), which may explain some punctual reconstruction discrepancies. However, IGAN managed to generate and reconstruct fairly plausible target signatures, including the variety of background levels, targets orientations and shadows.

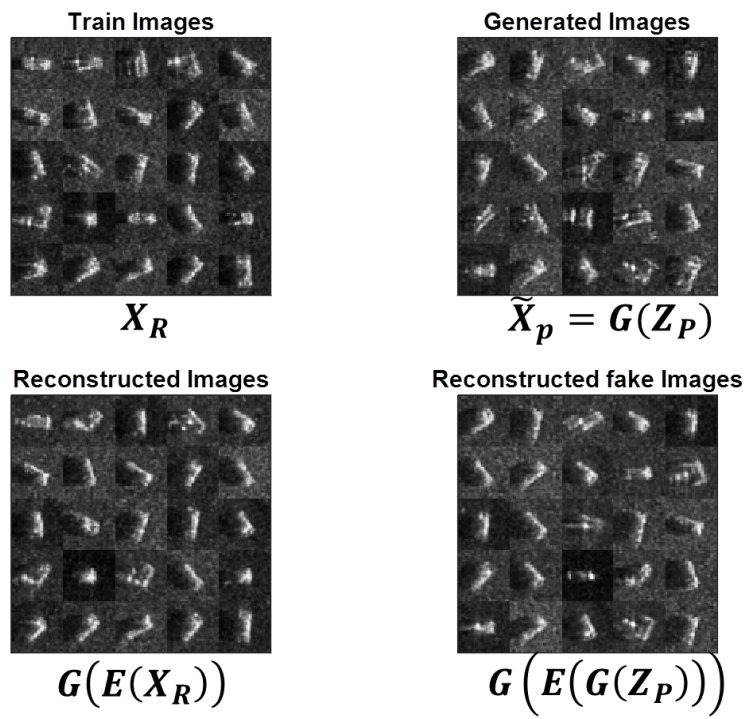

**Figure 2: SAMPLE dataset [24]: examples of SAR image chips used for IGAN training (up left), reconstruction of training images (down left), some images generated by IGAN (up right), and reconstruction of generated images (down right)**

The filtering effect of non-statistically relevant details and outliers is more obvious when IGAN is trained on the ANIME-FACE dataset [25] (over 60000 images of manga faces): for instance glasses, hairbands or fingers are not taken into account in the generation or reconstruction process as expected from their relative scarcity among the training dataset images (cf. Figure 3).

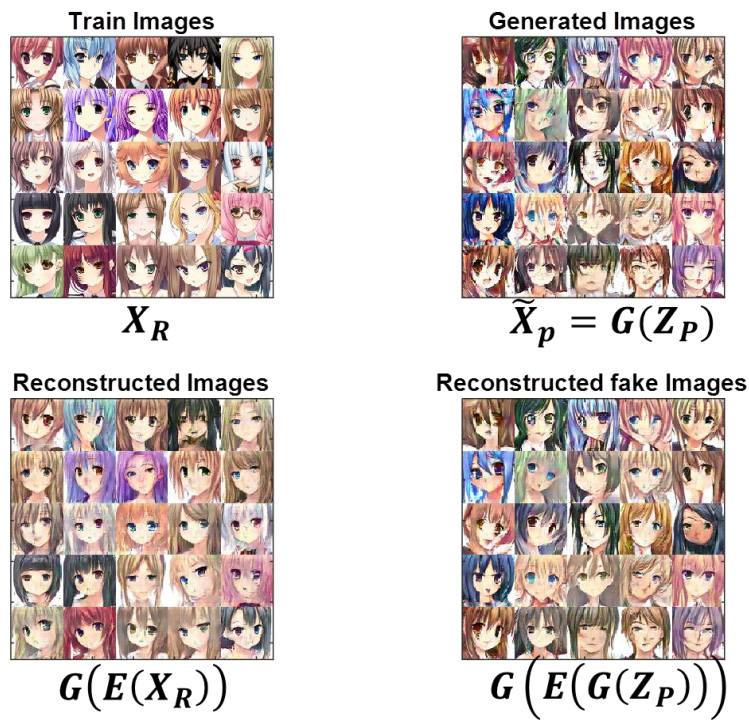

**Figure 3: auto-encoded reconstructions of real and generated images with IGAN trained on ANIME-FACE dataset [25]**

## 3.2 Self-supervised learning properties

IGAN was trained on the MNIST digits database [27] and the t-distributed Stochastic Neighbour Embedding (t-SNE) statistical method [26] was used to visualize the inferred latents distribution for Train and Test images (cf. Figure 4). Note that this training was done with no supervision, i.e. without any use of the digits label or even the number of classes. Figure 4 results show that latents belonging to the same digit class form relatively compact clusters on train set, and that we remarkably find the same clusters on full test set (simple and hard cases mixed). Some clusters boundaries are even clearly separated from others. This highlights the nice self-organisation property of the latent space generated by IGAN, with close latents corresponding to semantically close data. A fully un-supervised classification and class cardinality estimation would of course require an automatic clustering post-processing. Nevertheless, this example shows that IGAN presents some competitive properties with semi-supervised methods like InfoGAN [5], disentangling ones like BetaVAE [6], or clustering others like ClusterGAN [4].

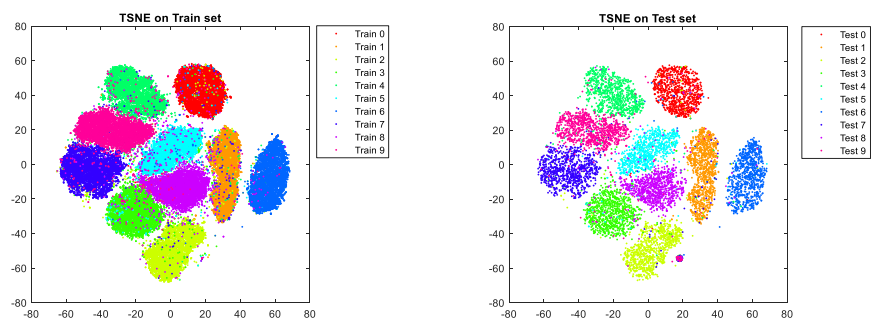

**Figure 4: IGAN trained on MNIST dataset [27]: t-SNE of latents generated on training set (left), and on test set (right), both with post-training colour visualisation of digits ground truth**

## 3.3 Latent space exploration and data transformations

The self-organization provided by IGAN enables us to easily compute some now classical interpolations and arithmetic in the latent space. As an illustration, Figure 5 presents the bijective transformation of "serious brunettes into smiling blondes and vice-versa" with IGAN being trained on the CelebA dataset [28]. The hair colour and smile attribute tags were used to select the desired image/latent couples, compute the mean latents corresponding to either characteristics, and do the substract/add arithmetic with the complementary mean latents to obtain the cross-characteristic images reconstructions. As expected, the transformation retains the facial personality and the head position of the original images, while filtering out the non-statistically relevant details/outliers (background, hat, scarf, arm, boyfriend ...).

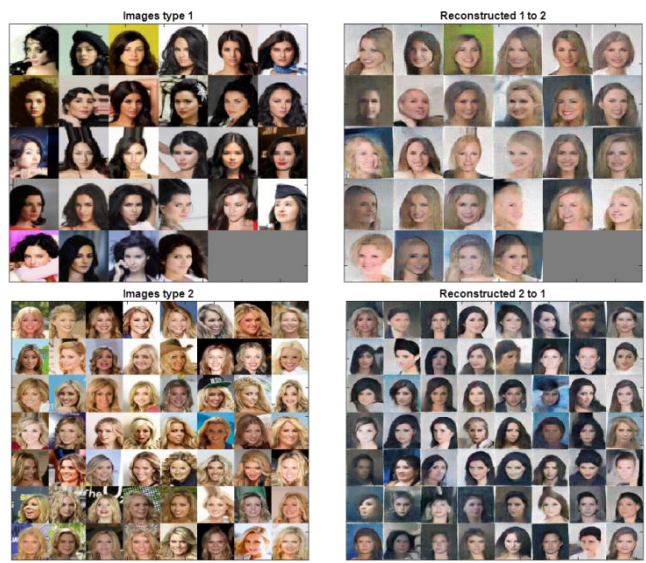

**Figure 5: CelebA [28] images with "serious & brunette" characteristics (up left), "smiling & blonde" (down left), IGAN reconstruction of "serious & brunette" changed into "smiling & blonde" (up right), and IGAN reconstruction of opposite change (down right)**

## 3.4 Shared latent space and multi domain transformations

Several publications have already explored the exploitation of a common or shared latent space, especially in the context of multi-modal image transformations, e.g. UNIT [29], MUNIT [30] or XGAN [31]. Most of the proposed solutions are limited to a dual image domain translation, whereas the IGAN architecture is *de facto* well suited for multiple data domains change (not limited to images) on a single shared latent space. Moreover, IGAN does not require a simultaneous training of paired domains, as each domain keeps its own separated Encoder & Generator training. Since IGAN spans the entirety of the latent space for every domain, any data in one domain will be turned into a plausible data for any other domain. Figure 6 illustrates the reconstruction of domain $A$ (ANIME-FACE [25]) and domain $B$ (CelebA [28]) images through successive cross-domain translations $\{G_A(E_B(G_B(E_A(A)))\}$ and $\{G_B(E_A(G_A(E_B(B)))\}$. Perfect reconstructions are not expected for data going through so many networks, but it is noticeable that the main image characteristics remain clearly unchanged. Domains that shares common semantic features should probably be merged into a single IGAN training to avoid unrelated latent correspondences between domains (e.g. non corresponding head orientations in both domains in Figure 6 example). Dedicated optical images manipulations often use separated latent spaces for style and content [32], and the IGAN framework could be advantageously adapted to this context in the future.

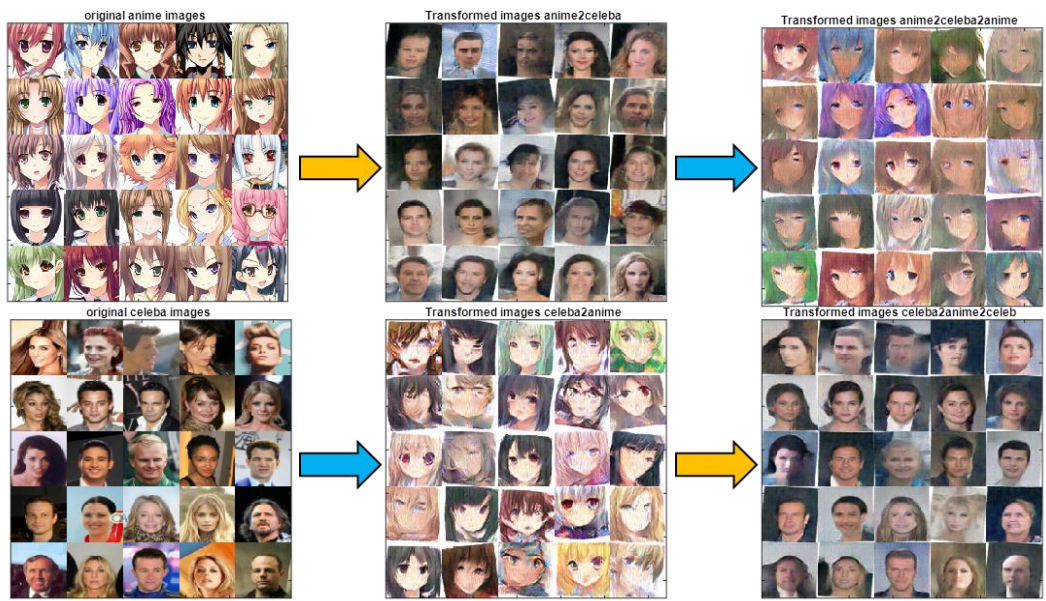

**Figure 6: reconstruction of ANIME-FACE [25] and CelebA [28] images through successive cross-domain (ANIME-FACE ⬅➡ CelebA) translations**

## CONCLUSIONS

IGAN builds a symmetric entangled adversarial game and a bijective auto-encoder in both data and latent spaces with multiple advantages:

- It balances, stabilizes the generation of data (and latents) and avoids mode collapse, while keeping the data generative quality. It provides measurable performance and convergence indicators.

- It spans generated latents on the full prior space, giving access to a wider generated space and thus to more variability of generated data.

- It provides an un-supervised self-organization of the latent space and enables the reconstruction of real and generated images and latents.

- It remains simple and frugal adding only an Encoder network to the Vanilla GAN.

This leads to numerous latent based applications for complex high dimensional data processing, like for instance semantic attributes arithmetic and interpolation, self-supervised learning, or multi domain translations. As a future work, a conditional version (CondIGAN) and fractionable latent space extensions are being considered. The original adversarial game developed for IGAN also suggests to revisit DiscoGAN and similar methods from a single Discriminator perspective using concatenated images from both domains as input.

## REFERENCES

[1] Goodfellow I., et al., 'Generative Adversarial Nets', Advances in Neural Information Processing Systems 27 NIPS. 2014

[2] Kingma D.P. and Welling M., 'Auto-Encoding Variational Bayes', 2013. https://arxiv.org/abs/1312.6114

[3] Kodali N., Abernethy J., Hays J., and Kira Z., 'On convergence and stability of GANs'.

arXiv:1705.07215, 2017.

[4] Mukherjee S., Asnani H., Lin E., and Kannan S., 'Cluster-GAN : Latent space clustering in generative adversarial net-works', CoRR, abs/1809.03627, 2018.

[5] Chen Duan Y., Houthooft R., Schulman J., Sutskever I. and Abbeel P., 'Infogan: Interpretable representation learning by information maximizing generative adversarial nets', in Neural Information Processing Systems, 2016, pp. 2172–2180.

[6] Higgins I., Matthey L., Pal A., Burgess C., Glorot X., Botvinick M., Mohamed S., Lerchner A., 'beta-VAE: Learning Basic Visual Concepts with a Constrained Variational Framework', ICLR 2017.

[7] https://towardsdatascience.com/understanding-latent-space-in-machine-learning-de5a7c687d8d

[8] Anders Boesen Lindbo Larsen, Søren Kaae Sønderby, Hugo Larochelle, and Ole Winther. 2016. Autoencoding beyond pixels using a learned similarity metric. In Proceedings of the 33rd International Conference on International Conference on Machine Learning - Volume 48 (ICML'16). JMLR.org, 1558–1566.

[9] Donahue J., et al. 'Adversarial Feature Learning', ICLR 2017. ArXiv abs/1605.09782

[10] Dumoulin, Vincent et al. 'Adversarially Learned Inference', ArXiv abs/1606.00704 2017.

[11] Li C., Liu H., Chen C., Pu Y., Chen L., Henao R., and Carin L., 'ALICE: Towards Understanding Adversarial Learning for Joint Distribution Matching'. In Advances in Neural Information Processing Systems, 2017.

[12] Ulyanov D., Vedaldi A., and Lempitsky V., 'It takes (only) two: Adversarial generator-encoder networks', in AAAI, 2018.

[13] Huang H. et al., 'IntroVAE: Introspective Variational Autoencoders for Photographic Image Synthesis', in NeurIPS 2018.

[14] Donahue J. and Simonyan K., 'Large Scale Adversarial Representation Learning', NeurIPS 2019.

[15] Lazarou C., 'Autoencoding Generative Adversarial Networks', ArXiv abs/2004.05472 2020.

[16] Li, Alexander Hanbo et al., 'Decomposed Adversarial Learned Inference', ArXiv abs/2004.10267 2020.

[17] Zhu, Jiapeng et al., 'Disentangled Inference for GANs with Latently Invertible Autoencoder', arXiv: Learning 2019.

[18] Pidhorskyi S., Adjeroh DA., Doretto G., 'Adversarial latent autoencoders', in Proceedings of the IEEE Conference on Computer Vision and Pattern Recognition (CVPR) 2020.

[19] Karras T., Laine S., Aila T., 'A style-based generator architecture for generative adversarial networks', arXiv:181204948 2018.

[20] Zhu, Jun-Yan et al. 'Unpaired Image-to-Image Translation Using Cycle-Consistent Adversarial Networks', 2017 IEEE International Conference on Computer Vision (ICCV), pp2242-2251, 2017.

[21] Kim, Taeksoo et al., 'Learning to Discover Cross-Domain Relations with Generative Adversarial

Networks', ICML 2017.

[22] Radford, Alec et al., 'Unsupervised Representation Learning with Deep Convolutional Generative Adversarial Networks', CoRR abs/1511.06434, 2016.

[23] Jiang, Yifan et al., 'TransGAN: Two Pure Transformers Can Make One Strong GAN, and That Can Scale Up', 2021.

[24] Lewis B., Scarnati T., Sudkamp E., Nehrbass J., Rosencrantz, S., Zelnio E., 'A SAR database for ATR development: Synthetic and Measured Paired Labeled Experiment (SAMPLE),' Proceedings of SPIE Algorithms for Synthetic Aperture Radar XXVI; Orlando, USA 2019.

[25] https://github.com/bchao1/Anime-Face-Dataset

[26] Van der Maaten L.J.P, and Hinton G.E., 'Visualizing High-Dimensional Data Using t-SNE', Journal of Machine Learning Research, vol. 9, novembre 2008, p. 2579–2605.

[27] 'THE MNIST DATABASE of handwritten digits". Yann LeCun, Courant Institute, NYU Corinna Cortes, Google Labs, New York Christopher J.C. Burges, Microsoft Research, Redmond.

[28] Liu, Ziwei et al., 'Deep Learning Face Attributes in the Wild', 2015 IEEE International Conference on Computer Vision (ICCV) 2015, pp. 3730-3738.

[29] Liu, Ming-Yu et al., 'Unsupervised Image-to-Image Translation Networks', NIPS 2017.

[30] Huang, Xun et al. 'Multimodal Unsupervised Image-to-Image Translation', ArXiv abs/1804.04732 2018.

[31] Royer A. et al., 'XGAN: Unsupervised Image-to-Image Translation for many-to-many Mappings', ArXiv abs/1711.05139 , 2020.

[32] Abdal R. et al., 'Image2StyleGAN: How to Embed Images Into the StyleGAN Latent Space?' 2019 IEEE/CVF International Conference on Computer Vision (ICCV) (2019): 4431-4440.